\documentclass[runningheads]{llncs}
\usepackage{graphicx}

\usepackage{tikz}
\usepackage{comment}
\usepackage{amsmath,amssymb} 
\usepackage{color}
\usepackage{marvosym}

\usepackage{array}
\usepackage{multirow}
\usepackage{colortbl}
\usepackage{cite}
\usepackage[ruled]{algorithm2e}
\usepackage{enumitem}
\usepackage{bm}
\usepackage{bbm}
\definecolor{mygray}{gray}{.92}
\makeatletter
\newcommand{\thickhline}{%
    \noalign {\ifnum 0=`}\fi \hrule height 1pt
    \futurelet \reserved@a \@xhline
}

\newcommand{\ie}{\textit{i}.\textit{e}.}
\newcommand{\eg}{\textit{e}.\textit{g}.}
\makeatother

\begin{document}
\pagestyle{headings}
\mainmatter
\def\ECCVSubNumber{3936}  

\title{\textit{ProposalContrast}: Unsupervised Pre-training for LiDAR-based 3D Object Detection} 

\titlerunning{\textit{ProposalContrast}}

%
\author{Junbo Yin \inst{1\dag} \and
Dingfu Zhou \inst{2,3} \and
Liangjun Zhang \inst{2,3} \and
Jin Fang \inst{2,3,4} \and \\
Cheng-Zhong Xu \inst{4} \and
{\Letter}Jianbing Shen \inst{4}  \and
{\Letter}Wenguan Wang \inst{5}
}
\authorrunning{J. Yin et al.}
%
\institute{$^1$School of Computer Science, Beijing Institute of Technology \quad
$^2$Baidu Research \\
\mbox{$^3$National Engineering Laboratory of Deep Learning Technology and Application, China} \\
\mbox{$^4$SKL-IOTSC, CIS, University of Macau \,
$^5$ReLER, AAII, University of Technology Sydney}
\\
\email{\{yinjunbocn,wenguanwang.ai\}@gmail.com}
\quad
\url{https://github.com/yinjunbo/ProposalContrast}
}

\maketitle

\begin{abstract}
Existing approaches for unsupervised point cloud pre-training are constrained to either scene-level or point/voxel-level instance discrimination. Scene-level methods tend to lose local details that are crucial for recognizing the road objects, while point/voxel-level methods inherently suffer from limited receptive field that is incapable of perceiving large objects or context environments. Considering region-level representations are more suitable for 3D object detection, we devise a new unsupervised point cloud pre-training framework, called \textit{ProposalContrast}, that learns robust 3D representations by contrasting region proposals. Specifically, with an exhaustive set of region proposals sampled from each point cloud, geometric point relations within each proposal are modeled for creating expressive proposal representations. To better accommodate 3D detection properties, \textit{ProposalContrast} optimizes with both inter-cluster and inter-proposal separation, \ie, sharpening the discriminativeness of proposal representations across semantic classes and object instances. The generalizability and transferability of \textit{ProposalContrast} are verified on various 3D detectors (\ie, PV-RCNN, CenterPoint, PointPillars and PointRCNN) and datasets (\ie, KITTI, Waymo and ONCE).

\keywords{3D Object Detection, Unsupervised Point Cloud Pre-training }
\end{abstract}

\let\thefootnote\relax\footnotetext{{\Letter}: Corresponding author. \quad {\dag}: Work done during an internship at Baidu Research.}

\begin{figure}[t]
{
\centering
\begin{center}
\includegraphics[width=0.98\textwidth]{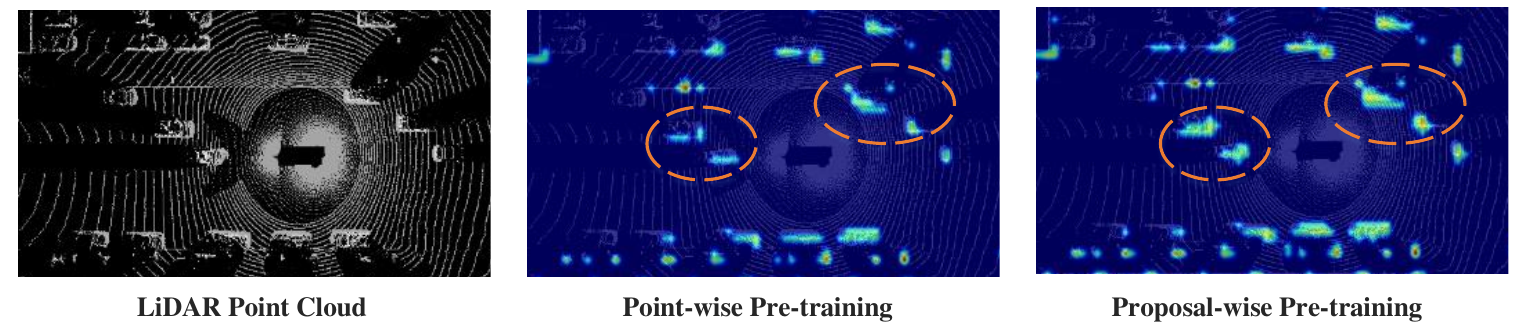}
\label{fig:features}
(a) \small{Compared with our proposal-level pre-training, point-level pre-training tends to produce incomplete object representations (as indicated by the dotted circle).}   \hfill\mbox{}
\end{center}
\centering
\begin{center}
\includegraphics[width=0.98\textwidth]{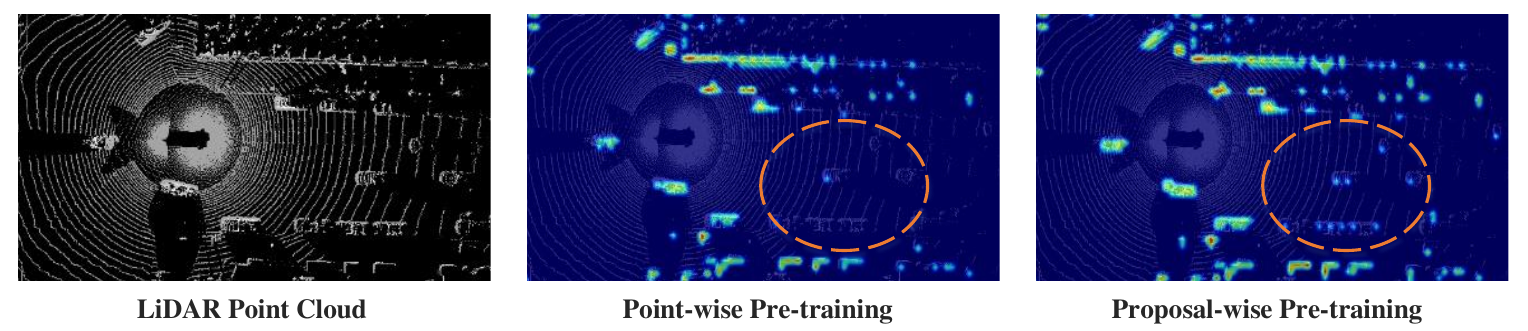}
\label{fig:fn}
(b) \small{By exploring larger context, our proposal-level pre-training successfully perceives objects that are missed by the point-wise method (as indicated by the dotted circle).}   \hfill\mbox{}
\end{center}

}%
\caption{\textbf{Comparison of VoxelNet representations} learned from PointContrast~\cite{xie2020pointcontrast} and our \textit{ProposalContrast}.}
\label{fig:qualitative}
\end{figure}

\section{Introduction}\label{sec:intro}

3D object detection from LiDAR point clouds has received great interest in recent years due to its significance to self-driving vehicles. Most existing 3D detectors are trained on massive labeled data. However, annotating point clouds is expensive and time-consuming. On the other hand, unlabeled point cloud data can be easily generated by self-driving vehicles: it is estimated that a self-driving car could collect 200k point cloud frames within only 8 working hours.

Self-supervised learning (SSL)~\cite{chen2020simple,he2020momentum,gidaris2020learning,zhai2019s,li2019joint} provides a feasible way to make use of unlabeled data. SSL methods typically define a {pretext task}, where free supervision signals can be derived from the data itself for representation learning. With a few labeled data of downstream tasks, the learnt representations can be fine-tuned and show excellent performance. Recent advances in SSL can be largely ascribed to contrastive learning~\cite{hjelm2018learning,chen2020simple,he2020momentum,chen2020exploring}. Contrastive learning explores the pretext task of instance discrimination, \ie, maximizing the agreement of the feature embeddings between two differently augmented views of the same data instance, and minimizing the agreement between different instances. The instances are often defined as images in the 2D domain. Intuitively, the instance discrimination pretext relies on an \textit{object-centric} assumption~\cite{chen2021multisiam, zhao2021self}, where the object of interest should lie in the center of an image such that different augmentations can be applied to achieve cross-view consistency. Although this is hold for some image datasets like ImageNet~\cite{deng2009imagenet}, it is suboptimal for LiDAR point cloud datasets; taking Waymo~\cite{sun2020scalability} as an example, objects are of smaller sizes (\eg, $4\times2m^{2}$ for car), and are unevenly distributed across a considerably wide range (\eg, $150\times150m^{2}$ for a point cloud scene).

Therefore, how to define the instances is crucial for the adoption of contrastive SSL techniques in point clouds. However, previous approaches for point cloud SSL either directly contrast different views of a whole point cloud scene~\cite{zhang2021self,huang2021spatio}, or merely focus on point-/voxel-level instance discrimination~\cite{xie2020pointcontrast,liang2021exploring}. The scene-level methods struggle to describe the locality of the road objects, while the point-/voxel-level methods overemphasize fine-grained details, lacking object-level characteristics. Each point cloud consists of several objects and the objects such as vehicles typically contain numerous points and span several voxels. Thus previous approaches take little consideration of the properties of point data, hurting the utility of the learned representations in downstream tasks, such as 3D object detection. In light of the analysis above, we believe learning point representations on the proposal-level is more desired (see Fig.~\ref{fig:features} for more evidence).

In this work, we propose a proposal-level point cloud SSL framework, named \textit{ProposalContrast}, that conducts proposal-wise contrastive pre-training for 3D detection-aligned representation learning. In particular, for each point cloud region proposal, its representation is designed to explicitly encode the geometrical relations of the points inside the proposal. This is achieved by a cross-attention based encoding module, which attends each point to its neighbors~\cite{liu2021swin}. To better align proposal-level pre-training with the nature of 3D detection, we propose to jointly optimize two pretext tasks, \ie, inter-proposal discrimination (IPD) and inter-cluster separation (ICS). IPD is for instance-discriminative representation learning. Through minimizing a proposal-wise contrastive loss, proposal representations are encouraged to gather instance-specific characteristics, eventually benefiting the localization of objects. Differently, ICS is for class-discriminative representation learning.  Since class label is not available during pre-training, ICS conducts cluster-based contrastive learning. It groups proposals into clusters and enforces consistency between cluster predictions (\ie, pseudo and soft class labels) of different views of each proposal. In this way, ICS encourages the proposal representations to abstract instance-invariant, common patterns, hence facilitating the semantic recognition of objects.

We comprehensively demonstrate the generalizability of our \textit{ProposalContrast} on prevalent 3D detection network architectures, \ie, PV-RCNN~\cite{shi2020pv}, CenterPoint~\cite{yin2020center}, PointPillars~\cite{lang2019pointpillars}, and PointRCNN~\cite{shi2019pointrcnn}, as well as empirically validate the transferability on three popular self-driving point cloud datasets, \ie, Waymo~\cite{sun2020scalability}, KITTI~\cite{geiger2012we}, and ONCE~\cite{mao2021one}. Moreover, \textit{ProposalContrast} can significantly save the annotation cost, \eg, with only half annotations, our pre-trained PV-RCNN outperforms the scratch model trained with full annotations.

\section{Related Work}\label{sec:related_work}

\noindent\textbf{3D Object Detection.} Existing solutions for 3D object detection can be broadly grouped into two classes in terms of point representations, \ie, grid-based and point-based. Specifically, grid-based methods~\cite{yan2018second,lang2019pointpillars,yin2020lidar,yin2021graph,zhou2020joint} typically first discretize the raw point clouds into regular girds (\eg, voxels~\cite{yan2018second} or pillars~\cite{lang2019pointpillars}), which can then be processed by 3D or 2D convolutional networks. Point-based methods~\cite{yang20203dssd,shi2019pointrcnn,meng2020weakly,meng2021towards,qi2017pointnet} directly extract features and predict 3D objects from raw point clouds. In general, grid-based approaches are efficient but suffer from information loss in the quantification process. Point-based methods yield impressive results, but the computational cost is high for large-scale point clouds. Hence some recent detectors~\cite{shi2020pv}, built upon the hybrid of voxel- and point-based architectures, are developed to enjoy the best of both worlds. In this work, we show that a wide range of modern 3D detectors can benefit from our self-supervised pre-training algorithm, which can learn meaningful and transferable representations from large-scale unlabeled LiDAR point clouds.

\noindent\textbf{Self-supervised Learning (SSL) in Point Cloud.} SSL~\cite{oord2018representation,jing2020self,hjelm2018learning,zhai2019s,kingma2014semi,ye2020probabilistic} is to learn expressive feature representation without manual annotations. Recently, contrastive learning based SSL algorithms~\cite{hjelm2018learning,chen2020simple,he2020momentum,wang2021exploring,yin2022semi} proved impressive results on various downstream tasks, even surpassing the supervised alternatives. In this article, we follow this paradigm with a proposal-level pretraining method specifically designed for the task of point cloud object detection.  Concurrent to our study, PointContrast~\cite{xie2020pointcontrast}, DepthContrast~\cite{zhang2021self}, GCC-3D~\cite{liang2021exploring}, and STRL~\cite{huang2021spatio} also exploit the potential of contrastive SSL in point cloud pretraining. Though impressive, these methods have a few limitations. First, \cite{zhang2021self,huang2021spatio} takes the whole point cloud scene as the instance, neglecting the underlying \textit{object-centric} assumption~\cite{chen2021multisiam, zhao2021self}, since self-driving point cloud scenes typically comprise of multiple object instances. Second, other methods like~\cite{xie2020pointcontrast,liang2021exploring} only take into account instance discrimination at the point-/voxel-level. Thus they are hard to acquire object-level representations that are compatible with 3D object detection. Third, \cite{xie2020pointcontrast,zhang2021self,huang2021spatio} overlook the semantic relations among instances, focusing on modeling low-level characteristics instead of high-level informative patterns. In \cite{liang2021exploring}, although an extra self-clustering strategy is introduced for capturing semantic properties, it only provides supervisory signals on moving voxels which are too sparse to cover potential object candidates, and leads to a complicated two-stage pipeline that trains the 3D encoder and 2D encoder separately. Differently, our point cloud pretraining method encourages discrimination between proposal instances and clusters, hence comprehensively capturing the properties of point cloud data and well aligning with 3D detection. Although RandomRooms~\cite{rao2021randomrooms} also considers region-level representations, it refers to synthetic CAD objects in indoor environments. In contrast,  we automatically mine potential object instances from raw point data in self-driving scenes.




\section{Approach}\label{sec:method}
In Sec.~\ref{subsec:overall}, we first briefly describe our proposal-level pre-training method specifically designed for 3D object detection. Then, we detail the crucial components, \ie, region proposal encoding module (in Sec.~\ref{subsec:rpe}) and joint optimization of inter-proposal discrimination and inter-cluster separation (in Sec.~\ref{subsec:CID}).

\begin{figure*}[t]
  \centering
      \includegraphics[width=0.99\linewidth]{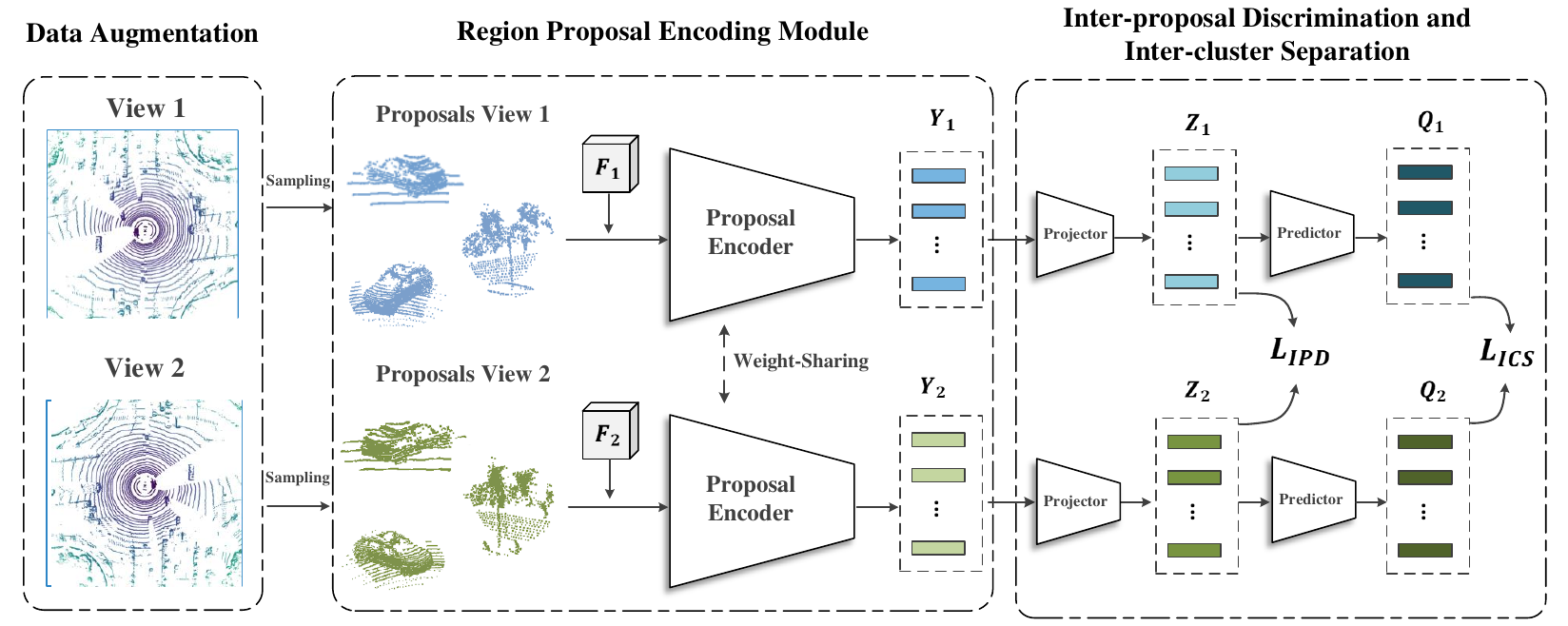}
\caption{\textbf{Illustration of our \textit{ProposalContrast} framework}. Given augmented point cloud with different views, we first sample paired region proposals and then extract the features with a region proposal encoding module. After that, inter-proposal discrimination and inter-cluster separation are enforced to optimize the whole network.}
\label{fig:framework}
\end{figure*}

\subsection{Overview of \textit{ProposalContrast}}
\label{subsec:overall}
Due to the specific characteristics of 3D LiDAR point cloud data, such as the irregular and sparse structures across large perception ranges, simply applying 2D pre-training techniques to point clouds cannot get satisfactory results. This calls for better adapting existing pre-training techniques to the inherent structures of point cloud data. Rather than current point cloud pretraining methods investigating the unsupervised representation learning on the scene-/point-/voxel-level~\cite{xie2020pointcontrast,zhang2021self,huang2021spatio,xie2020pointcontrast}, our \textit{ProposalContrast} learns representations by contrasting directly on region proposals.

As shown in Fig.~\ref{fig:framework}, to achieve proposal-level pre-training, \textit{ProposalContrast} has five core components: data augmentation, correspondence mapping, region proposal generation, region proposal encoding, as well as a joint optimization of inter-proposal discrimination (IPD) and inter-cluster separation (ICS).

\noindent\textbf{Data Augmentation.} Let ${X}_0\!\in\!\mathbb{R}^{L_0\times{3}\!}$ denote an input point cloud with~$L_0$~points (here we describe point cloud with 3D coordinates for simplicity). We apply two different data augmentation operators $\mathcal{T}_1, \mathcal{T}_2$ on ${X}_0$ to produce two augmented views ${X}_{1}$, ${X}_{2}$:
\begin{equation}
{{X}_{1}} = \mathcal{T}_1({X}_0)\!\in\!\mathbb{R}^{L_{1}\times{3}},~~~~{{X}_{2}}=\mathcal{T}_2({X}_0)\!\in\!\mathbb{R}^{L_{2}\times{3}},
\end{equation}
where $L_1$ and $L_2$ are number of points of ${X}_{1}$ and ${X}_{2}$. The data augmentation strategies include random rotation, flip, scaling and random point drop out (see Sec.~\ref{sec:experiments} for detailed definition). Note that random rotation is only applied on the upright axis since we are aware of self-driving scenarios.

\noindent\textbf{Correspondence Mapping.} Before generating proposals, we first get the correspondence mapping ${{M}}$ between point sets in ${X}_{1}$ and ${X}_{2}$. This can be easily achieved by recording the index of each point in the original view ${X}_0$. ${{M}}$ is later used for sampling and grouping region proposals.

\noindent\textbf{Region Proposal Generation.} Some SSL methods for 2D representation learning make use of image proposals in the form of 2D bounding boxes~\cite{chen2021multisiam}. For point clouds, straightforwardly representing proposals as 3D bounding boxes, however, is not a feasible choice, due to the significantly enlarged space of object candidates in 3D scenarios. In addition, generating a dense set of 3D bounding boxes will lead to unacceptable high computational cost.  These considerations motivate us to adopt spherical proposals, instead of 3D bounding box proposals. Specifically, given the input point cloud ${X}_0$, we first abandon the road plane points so as to sample less from the background~\cite{bogoslavskyi2017efficient}. Then we perform farthest point sampling (FPS)~\cite{qi2017pointnet} on ${X}_0$ to sample a total of $N$ points as the centers of $N$ spherical proposals (the corresponding samples in ${X}_{1}$ and ${X}_{2}$ can also be identified according to the correspondence ${{M}}$). FPS encourages the sampled points to be away from each other and thus guarantees the diversity of the sampled proposals. Next, spherical proposals are generated by searching $K$ points around each sampled center point within a pre-defined radius $r$. Finally, we get two sets of spherical proposals ${P}_{1}, {P}_{2}\!\in\!\mathbb{R}^{N\times{K}\times{3}}$ from the two views, \ie, ${P}_{1}\in{X}_{1}, {P}_{2}\in{X}_{2}$.

\noindent\textbf{Region Proposal Encoding.} An encoding module is further adopted to extract expressive proposal representations, by considering the geometric relations of the points inside each proposal. Given the scene-level representations, \ie, $\bm{F}_1$ and $\bm{F}_2$, of ${X}_1$ and ${X}_2$, extracted by the backbone network, the proposal encoding module outputs geometry-aware proposal representations $\bm{Y}_{1}, \bm{Y}_{2}\!\in\!\mathbb{R}^{N\times{C}}$ for ${P}_1$ and ${P}_2$:
\begin{equation}
{{\bm{Y}_{1}} =f_{\text{En}}({{P}_{1}}, \bm{F}_1)},~~~{\bm{Y}_{2}} =f_{\text{En}}({{P}_{2}},  \bm{F}_2),
\end{equation}
where the encoding module $f_{\text{En}}$ is achieved by a neural attention mechanism, which will be detailed in Sec.~\ref{subsec:rpe}.

\noindent\textbf{Joint Optimization.} The proposal features $\bm{Y}_{1}$, $\bm{Y}_{2}$ from different views are learned by enforcing both cluster-based class consistency and instance-wise discrimination. This endows $\bm{Y}_{1}$, $\bm{Y}_{2}$ with desired properties for 3D object detection. The detailed design is presented in Sec.~\ref{subsec:CID}.


\begin{figure}[t]
\begin{center}
\includegraphics[width=0.75\textwidth]{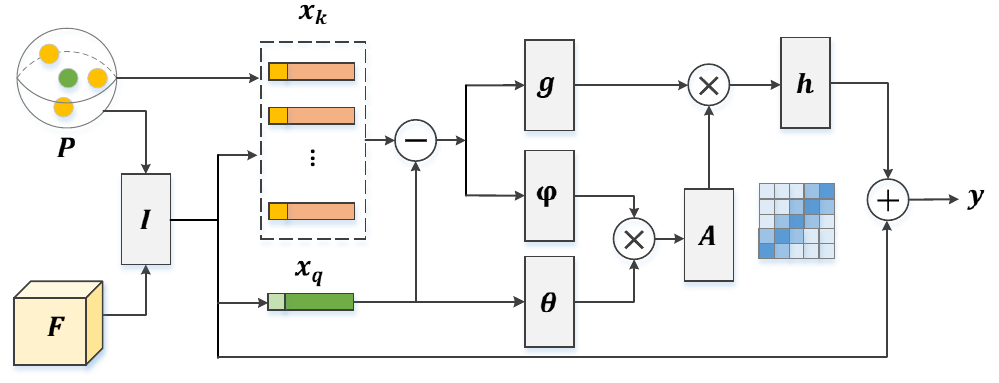}
\end{center}
\caption{ \small{\textbf{Illustration of the region proposal encoding module} (Sec.~\ref{subsec:rpe}). The encoding module adopts a cross-attention mechanism that attends the center query point with its neighbor key points, for collecting expressive proposal representations.}}
\label{figs:encoding}
\end{figure}

\subsection{Region Proposal Encoding Module}
\label{subsec:rpe}
The relations among points in a proposal provide crucial geometry information for describing the proposal. To extract better proposal representations, we leverage the cross-attention~\cite{dosovitskiy2020image} to capture the geometric relations between points.

For a point cloud scene ${X}$ and corresponding proposals ${P}$, we extract its global scene-wise representation through a backbone network, \ie, $\bm{F}\!=\!f_{\text{Bbone}}({{X}})$. Popular 3D backbones like VoxelNet~\cite{yan2018second} and PointNet++~\cite{qi2017pointnet} can be used to instantiate $f_{\text{Bbone}}$.

Then we obtain initial representation ${\bm{P}}\!\in\!\mathbb{R}^{{N}\times{K}\times{C}}$ for all the proposals ${{P}}$ by applying bilinear interpolation function $I(\cdot)$ over $\bm{F}$, \ie, ${\bm{P}}=I({{P}}, \bm{{F}})$, where $N$ is the proposal number of one view, $K$ is the point number inside a proposal and $C$ is the backbone channel number.

After that, we capture the geometrical relation between points via cross attention, \ie, attending a query point with other key points.  Let ${\bm{p}}\!\in\!{\bm{P}}$ with ${K}\times{C}$ size denote the initial representation of a proposal ${{p}}\!\in\!{{P}}$. As illustrated in Fig.~\ref{figs:encoding}, we set the center point feature ${\bm{x}_q}\!\in\!\mathbb{R}^{{1}\times{C}}$ of proposal $p$, \ie, ${\bm{x}_q}\!\in\!\bm{p}$ as the query, since the center point is more informative~\cite{yin2020center}. Next, we get the neighbor features ${\bm{x}_k}\!\in\!\mathbb{R}^{{K}\times{C}}$, where ${\bm{x}_k}\!\in\!\bm{p}$, and use the difference between ${\bm{x}_k}$ and ${\bm{x}_q}$ as keys to encode the asymmetric geometry relation as recommended by~\cite{wang2019dynamic}. The point coordinate of $\bm{x}_q$ and $ \bm{x}_k$ are also integrated to provide explicit position information.

Formally, the ${\bm{x}_q}$ and ${\bm{x}_k}$ are first projected to query, key, and value vectors through:
\begin{equation}\label{eq:3}
{\bm{{w}}_q \!=\!\theta({\bm{{x}}_q})},~~~
{\bm{{w}}_k\!=\!\phi({\bm{{x}}_k}\!-\!{\bm{{x}}_q})},~~~
{\bm{{w}}_v \!=\!g({\bm{{x}}_k}\!-\!{\bm{{x}}_q})},
\end{equation}
where $\theta(\cdot)$, $\phi(\cdot)$, and $g(\cdot)$ are linear layers. Afterwards, we compute the attention weight $A\in[0,1]^K$ based on normalized similarities between the query $\bm{w}_q$ and each key $\bm{w}_k$. The attention weight is then applied on the value vector ${\bm{w}_v}$ to aggregate information from all the keys:
\begin{equation}
\begin{aligned}
\bm{{w}}_{o} = \sum\nolimits_{\bm{w}_k}A({\bm{{w}}_q},\bm{w}_k)\cdot{\bm{w}_v},~~A({\bm{{w}}_q},\bm{w}_k)=\frac{{\bm{{w}}^\top_q}\bm{w}_k}{\sum_{\bm{w}_k}{\bm{{w}}^\top_q}\bm{w}_k}.
\end{aligned}
\end{equation}
As seen, the attention weight encodes the geometry relation of each point pair under a metric space, which makes the center point aware of the informative neighborhoods and thus encourages effective information exchange between points inside a proposal. Finally, the representation of proposal $p$ is given as:
\begin{equation}\label{eq:5}
\bm{{y}}=\bm{{x}}_q+h(\bm{{w}}_{o}),
\end{equation}
where $h(\cdot)$ is a linear layer. In this way, by applying Eqs.~(\ref{eq:3}-\ref{eq:5}) on each the region proposal in ${{X}_1}$ and ${{X}_2}$, we got all the proposal representations, \ie, ${\bm{Y}_1}, {\bm{Y}_2}\in\mathbb{R}^{{N}\times{C}}$, which will be trained by  joint optimization of inter-proposal discrimination and inter-cluster separation.

\subsection{Joint Optimization of Inter-proposal Discrimination and Inter-cluster Separation}
\label{subsec:CID}
A good point cloud representation for 3D detection is desired to be discriminative across both instances and classes; object instance-sensitive representation benefits the localization of 3D objects, while class-discriminative representation is crucial for the recognition of object categories. We therefore propose to optimize with both inter-proposal discrimination (IPD) and inter-cluster separation (ICS) simultaneously. Because annotations for object bounding box and semantics labels are not given during pretraining, IPD is designed to contrast the representations on the proposal-level, while ICS discovers stable and informative 3D patterns through grouping the proposals into clusters.

\noindent\textbf{Inter-proposal discrimination (IPD).} For object instance-sensitive representation learning, we conduct proposal-level contrastive learning~\cite{chen2020simple}. Specifically, given the proposal representations, \ie, ${\bm{Y}_1}, {\bm{Y}_2}\!\in\!\mathbb{R}^{{N}\times{C}}$, from the two different views $X_1, X_2$. We aim to pull positive pairs (\eg, the same proposal with different views) close, as well as push negative pairs (\eg, the different proposals) apart.

Following the common practice in contrastive learning, we first adopt a projection layer $f_\text{Proj}(\cdot)$ to project each proposal representation $\bm{y}\!\in\!\{\bm{Y}_1, \bm{Y}_2\}$ to a $\ell_2$-normalized embedding space:
\begin{equation}
\bm{z}=\frac{f_\text{Proj}(\bm{y})}{||f_\text{Proj}(\bm{y})||_2} \in [0,1]^C.
\end{equation}
Given the sets of normalized proposal embeddings $\bm{Z}_1, \bm{Z}_2\!\in\![0,1]^{{N}\times{C}}$, the IPD loss is designed in a form of the InfoNCE loss~\cite{oord2018representation}:
\begin{equation}\label{eq:7}
\mathcal{L}_\text{IPD}\!=\!\frac{1}{N}\!\!\!\sum_{\bm{z}^n_1\in\bm{Z}_1}\!\!\!-\!\log{\frac{\exp(\bm{z}^{n\top\!}_{1}\!\!\cdot\!\bm{z}^{n'}_{2}/\tau)}{\!\!\!\sum\limits_{\bm{z}^m_2\in\bm{Z}_2}\!\!\!\exp({\bm{z}^{n\top\!}_{1}\!\!\cdot\!\bm{z}^{m}_{2}/\tau})}}
+\frac{1}{N}\!\!\!\sum_{\bm{z}^m_2\in\bm{Z}_2}\!\!\!-\!\log{\frac{\exp(\bm{z}^{m\top\!}_{2}\!\!\cdot\!\bm{z}^{m'}_{1}/\tau)}{\!\!\!\sum\limits_{\bm{z}^n_1\in\bm{Z}_1}\!\!\!\exp({\bm{z}^{m\top\!}_{2}\!\!\cdot\!\bm{z}^n_1/\tau})}} ,
\end{equation}
where $\tau$ is a temperature hyper-parameter. For $\bm{z}^{n}_{1}$, $\bm{z}^{n'}_{2}$ refers to a positive sample; $\bm{z}^{n}_{1}$ and  $\bm{z}^{n'}_{2}$ correspond to a same proposal in $X_0$. Similarly, $\bm{z}^{m}_{2}$ is a negative sample for $\bm{z}^{n}_{1}$, \ie, $\bm{z}^{m}_{2}\!\neq\!\bm{z}^{n'}_{2}$.

\noindent\textbf{Inter-cluster Separation (ICS).} For class-discriminative representation learning, we group the spherical proposals into different clusters, which can be viewed as pseudo class labels. Inspired by recent clustering-based contrastive SSL~\cite{caron2020unsupervised}, ICS is designed to maximize the agreement between cross-view cluster assignments.

For each normalized proposal feature $\bm{{z}}^n_1\!\in\!\bm{Z}_1$ from view $X_1$, we first apply a predictor $f_\text{Pred}(\cdot)$ to map it to vector $\bm{{q}}^n_1\!\in\!\mathbb{R}^O$ that represents its pseudo-class embedding, such that $\bm{{q}}^n_1\!=\!f_\text{Pred}(\bm{{z}}^n_1)$, where the output channel dimension $O$ of $\bm{{q}}^n_1$ refers to the number of pseudo classes.  To get cluster assignment, \ie, $\hat{\bm{q}}^n_1\!\in\![0,1]^O$, we stop gradient on $\bm{{q}}^n_1$ and adopt Sinkhorn-Knopp algorithm~\cite{cuturi2013sinkhorn} to group all the proposals in each training batch into $O$ clusters. Then the training target of ICS is defined as:
\begin{equation}
\mathcal{L}_\text{ICS} = \frac{1}{N}\sum\nolimits_{\bm{q}^n_1}\!\!\!-\hat{\bm{q}}^n_1\log{\sigma(\bm{{q}}^{n'}_2)}+ \frac{1}{N}\sum\nolimits_{\bm{q}^m_2}\!\!\!-\hat{\bm{q}}^m_2\log{\sigma(\bm{{q}}^{m'}_1)},
\label{eq:pairloss}
\end{equation}
where $\sigma(\cdot)$ refers to the softmax function that maps the pseudo-class embedding into class probability distribution. $\mathcal{L}_\text{ICS}$ encourages the cross-view clustering consistency of each proposal, \ie, use the cluster assignment $\hat{\bm{q}}^n_1$ (resp. $\hat{\bm{q}}^m_2$) of view $X_1$ (resp. $X_2$) as pseudo grouptruth to supervise the class probability distribution $\sigma(\bm{{q}}^{n'}_2)$ (resp. $\sigma(\bm{{q}}^{m'}_1)$). Note that $\hat{\bm{q}}^n_1$ and $\bm{{q}}^{n'}_2$ correspond to a same proposal in $X_0$.

Finally, the overall self-supervised learning loss is defined as:
\begin{equation}\label{eq:totalloss}
\mathcal{L}=\alpha\mathcal{L}_\text{IPD}+\beta\mathcal{L}_\text{ICS},
\end{equation}
where $\alpha$ and $\beta$ are the balancing coefficients, respectively. In Sec.~\ref{sec:ablation}, we provide analysis on the effectiveness of the two optimization targets.



\section{Experimental Results}
\subsection{Pre-training Settings}
\label{sec:experiments}

\noindent\textbf{Datasets.} We adopt the common experimental protocol of SSL, \ie, first pre-training a backbone network with large-scale unlabeled data and then fine-tuning it on downstream tasks with much fewer labeled data. Some previous 3D SSL methods make use of ShapeNet~\cite{chang2015shapenet} and ScanNet~\cite{dai2017scannet} datasets to pre-train the 3D backbones, thus they only focus on the indoor setting and suffer from large domain gap when transferring to the self-driving setting. In our experiments, we adopt Waymo Open Dataset~\cite{sun2020scalability} for the self-supervised pre-training. The Waymo dataset contains 798 scenes (158,361 frames) for training and 202 scenes (40,077 frames) for validation; it is 20$\times$ larger than KITTI~\cite{geiger2012we}. We adopt the whole training set to pre-train various 3D backbones without using the labels.


\noindent\textbf{Network Architectures.} For thorough examination of the efficacy and versatility of our approach, we investigate the performance of \textit{ProposalContrast} on diverse 3D backbone architectures, including grid-based, \ie, VoxelNet~\cite{yan2018second} and PointPillars~\cite{lang2019pointpillars}, as well as point-based, \ie, PointNet++~\cite{qi2017pointnet}.  The projection layer, $f_\text{Proj}(\cdot)$, in Sec.~\ref{subsec:CID} is implemented as two linear layers, with the first layer followed by a batch normalization (BN) layer and a ReLU. The channel dimension of the output is set as $C\!=\!128$. The predictor $f_\text{Pred}(\cdot)$, implemented as a linear layer, outputs a 128-$d$ vector as the pseudo-class embedding, \ie, $O\!=\!128$. All the functions in the proposal encoding module are instantiated by linear layers with 128 channels. The attention linear head $h(\cdot)$ transforms the attended features to the original backbone channels.

\noindent\textbf{Implementation Details.} We empirically consider four types of data augmentations to generate different views, including random rotation ($[-180^\circ, 180^\circ]$), random scaling ([0.8, 1.2]), random flipping along X-axis or Y-axis, and random point drop out. For random point drop out, we sample 100k points from the original point cloud for each of the two augmented views.
%
20k points are chosen from the same indexes to ensure a 20\% overlap for the two augmented views,
while the other 80k points are randomly sampled from the remained point clouds.
We sample $N\!=\!2048$ spherical proposals for every point cloud frame; each proposal contains $K\!=\!16$ points within $r\!=\!1.0$  m radius.
The parameters for the VoxelNet~\cite{yan2018second} backbone are the same as the corresponding 3D object detectors.  The temperature parameter $\tau$ in the IPD loss $\mathcal{L}_\text{IPD}$ (Eq.~\ref{eq:7}) is set to 0.1. The coefficients $\alpha$ and $\gamma$ in Eq.~\ref{eq:totalloss} are both set to 1 empirically. We pre-train the models for 36 epochs, and use Adam optimizer~\cite{kingma2015adam} to optimize the network. Cosine learning rate schedule~\cite{loshchilov2016sgdr} is adopted with warmup strategy in the first 5 epochs.  The maximum learning rate is set to 0.003.

\subsection{Transfer Learning Settings and Results}
\label{sec:mAP}
In this paper, we investigate self-supervised pre-training in an autonomous driving setting. We evaluate our approach on several widely used LiDAR point cloud datasets, \ie, KITTI~\cite{geiger2012we}, Waymo~\cite{sun2020scalability} and ONCE~\cite{mao2021one}. Specifically, we compare our \textit{ProposalContrast} with other pre-training methods by fine-tuning the detection models. Different amounts of labeled data are used for fine-tuning to show the data-efficient ability. Besides, various modern 3D object detectors are involved to demonstrate the generalizability of our pre-trained models.

\begin{table}[t]
\caption{{\textbf{Data-efficient 3D Object Detection on KITTI.} We pre-train the backbones of PointRCNN~\cite{shi2019pointrcnn} and PV-RCNN~\cite{shi2020pv} on Waymo and transfer to KITTI 3D object detection with different label configurations. Consistent improvements are obtained under each setting. Our approach outperforms all the concurrent self-supervised learning methods, \ie, DepthContrast~\cite{zhang2021self}, PointContrast~\cite{xie2020pointcontrast}, GCC-3D~\cite{liang2021exploring}, and STRL~\cite{huang2021spatio}.}}
\label{tb:kitti}
\centering
\renewcommand\arraystretch{1.1}
\resizebox{1\textwidth}{!}
{
\begin{tabular}{c||c|c||c|ccc|ccc|ccc}
\hline\thickhline
		\rowcolor{mygray}
\textbf{Fine-tuning with}  && \textbf{Pre-train.} &\textbf{mAP}       &  \multicolumn{3}{c|}{\textbf{Car}}          & \multicolumn{3}{c|}{\textbf{Pedestrian}   }         & \multicolumn{3}{c}{\textbf{Cyclist}}         \\
		\rowcolor{mygray}
{\textbf{various label ratios}} &\multirow{-2}{*}{\textbf{Detector}} &\textbf{Schedule}
&\textbf{(Mod.)}  & \textbf{Easy}             & \textbf{Mod.}           & \textbf{Hard}         & \textbf{Easy}             & \textbf{Mod.}           & \textbf{Hard} &\textbf{Easy}             & \textbf{Mod.}           & \textbf{Hard}        \\ \hline\hline
  &  & Scratch & 63.51 & 88.64	&75.23	&72.47       & 55.49 	& 48.90 &	42.23    &85.41 &	66.39 	&61.74  \\
\multirow{2}{*}{20\% ($\sim$ 0.7k frames)}  &\multirow{-2}{*}{PointRCNN} & Ours&{\textbf{66.20}}$_\textbf{{+2.69}}$ & 88.52	&\textbf{77.02$_\textbf{{+1.79}}$}	&72.56 & 58.66 	&\textbf{51.90$_\textbf{{+3.00}}$} 	&44.98           &90.27 	&\textbf{69.67$_\textbf{{+3.28}}$} 	&65.05     \\ \cline{2-13}
&&  Scratch & 66.71 &91.81	&82.52	&80.11           &58.78 	&53.33 	&47.61   &86.74 	&64.28 	&59.53            \\
  & \multirow{-2}{*}{~~PV-RCNN} & Ours&{\textbf{68.13}}$_\textbf{{+1.42}}$&91.96	&\textbf{82.65$_\textbf{{+0.13}}$}	&80.15&62.58 &	\textbf{55.05$_\textbf{{+1.72}}$} &	50.06           &88.58 	&\textbf{66.68$_\textbf{{+2.40}}$} 	&62.32   \\ \cline{2-12}   \hline

 &     &Scratch  & 66.73 & 89.12	&77.85	&75.36          &61.82 	&\textbf{54.58} 	&47.90   &86.30 	&67.76 	&63.26   \\
\multirow{2}{*}{50\% ($\sim$ 1.8k frames)}  &\multirow{-2}{*}{PointRCNN} & Ours&{\textbf{69.23}}$_\textbf{{+2.50}}$&89.32	&\textbf{79.97$_\textbf{{+2.12}}$}	&77.39  & 62.19 	&{54.47$_\textbf{{-0.11}}$} 	&46.49          &92.26 	&\textbf{73.25$_\textbf{{+5.69}}$} 	&68.51   \\ \cline{2-13}
& & Scratch &69.63  & 91.77	 &82.68	&81.9           &63.70 	&57.10 	&52.77   &89.77 	&69.12 &	64.61           \\
  & \multirow{-2}{*}{~~PV-RCNN} &Ours &{\textbf{71.76}}$_\textbf{{+2.13}}$ &92.29	&\textbf{82.92$_\textbf{{+0.24}}$}	&82.09& 65.82 	&\textbf{59.92$_\textbf{{+2.82}}$} 	&55.06        &91.87 	&\textbf{72.45$_\textbf{{+3.33}}$} 	&67.53     \\   \hline

   &    & Scratch & 69.45 & 90.02	&\textbf{80.56}	&78.02        &62.59 	&55.66 	&48.69   &89.87 	&72.12 	&67.52   \\
   &    & DepthCon.~\cite{zhang2021self} & 70.26$_\textbf{{+0.81}}$ &89.38	&80.32$_\textbf{{-0.24}}$	&77.92          &65.55 	&57.62$_\textbf{{+1.96}}$ 	&50.98  &90.52 	&72.84$_\textbf{{+0.72}}$ 	&68.22   \\
\multirow{4}{*}{100\% ($\sim$ 3.7k frames)}  &\multirow{-3}{*}{PointRCNN}  & Ours&{\textbf{70.71}}$_\textbf{{+1.26}}$&89.51	&80.23$_\textbf{{-0.33}}$	&77.96  & 66.15 	&\textbf{58.82$_\textbf{{+3.16}}$} 	&52.00            &91.28 	&\textbf{73.08$_\textbf{{+0.96}}$} 	&68.45  \\ \cline{2-13}
& &Scratch  & 70.57 & -	&84.50	&-          &- 	&57.06	&-   &- 	&70.14 	&-           \\
& &GCC-3D~\cite{liang2021exploring}  & 71.26$_\textbf{{+0.69}}$ & -           & -        & -          &-   & -           &-  &-    &-           &-         \\
& &STRL~\cite{huang2021spatio}  & 71.46$_\textbf{{+0.89}}$ & -          & 84.70$_\textbf{{+0.20}}$        & -          &-    & 57.80$_\textbf{{+0.74}}$            &-  &-    & 71.88$_\textbf{{+1.74}}$            &-           \\
   &    & PointCon.~\cite{xie2020pointcontrast} & 71.55$_\textbf{{+0.98}}$ &91.40	&84.18$_\textbf{{-0.32}}$	&82.25          &65.73 	&57.74$_\textbf{{+0.68}}$ 	&52.46  &91.47 	&72.72$_\textbf{{+2.58}}$ 	&67.95   \\
  & \multirow{-5}{*}{~~PV-RCNN}. &Ours &{\textbf{72.92}}$_\textbf{{+2.35}}$&92.45	&\textbf{84.72$_\textbf{{+0.22}}$}	&82.47  & 68.43 	&\textbf{60.36$_\textbf{{+3.30}}$} 	&55.01           &92.77 	&\textbf{73.69$_\textbf{{+3.55}}$} 	&69.51      \\  \hline

\end{tabular}
}
\end{table}

\noindent\textbf{KITTI Dataset.} KITTI 3D object benchmark~\cite{geiger2012we} has been widely used for 3D object detection from LiDAR point cloud. It contains 7,481 labeled samples, which are divided into two groups, \ie, a training set (3,712 samples) and a validation set (3,769 samples). Mean Average Precision (mAP)  with 40 recall positions are usually adopted to evaluate the detection performance, with a 3D IoU thresholds of 0.7 for cars and 0.5 for pedestrians and cyclists.

We assess the transferability of our pre-trained model by pre-training on Waymo then fine-tuning on KITTI. Two typical 3D object detectors, \ie, PointRCNN~\cite{shi2019pointrcnn} and PV-RCNN~\cite{shi2020pv}, are used as the baselines. The two detectors are based on different 3D backbones (\ie, point-wise or voxel-wise networks), covering most cases of mainstream 3D detectors. A crucial advantage of self-supervised pre-training is to improve data efficiency for the downstream task with limited annotated data. To this end, we evaluate the data-efficient 3D object detection. In particular, the training samples are split into three groups, with each containing 20\% (0.7k), 50\% (1.8k) and 100\% (3.7k) labeled samples, respectively.
The experimental results are shown in Table~\ref{tb:kitti}. On both 3D detectors, our self-supervised pre-trained model effectively improves the performance in comparison to the model trained from scratch. Our model also outperforms several concurrent works. For example, based on PV-RCNN detector, our model exceeds  STRL~\cite{huang2021spatio} and GCC-3D~\cite{liang2021exploring} by 1.46\% and 1.66\%, respectively. Our model also surpasses DepthContrast~\cite{zhang2021self} and PointContrast~\cite{xie2020pointcontrast}, thanks to the proposal-level representation.
Besides, the classes with fewer labeled instances (\eg, pedestrian and cyclist) are improved a lot, showcasing the ability to address imbalanced class distribution.
More importantly, with our pre-trained model, PointRCNN and PV-RCNN using half annotation achieve comparative performance compared with the counterparts with full annotations. This also suggests the potential of our approach in reducing the heavy annotation burden.


\begin{table}[t]
\caption{{\textbf{Comparisons between our model and other self-supervised learning methods on Waymo}. All the detectors are trained by 20\% training samples following the OpenPCDet~\cite{openpcdet2020} configuration and evaluated on the validation set. Both PV-RCNN~\cite{shi2020pv} and CenterPoint~\cite{yin2020center} are used as beseline detectors.}}
\label{tb:waymo-D5}
\centering
\setlength\tabcolsep{2pt}
\renewcommand\arraystretch{1.1}
\resizebox{0.99\textwidth}{!}{
\begin{tabular}{c||c||c|ccc}
\hline\thickhline
\rowcolor[HTML]{EFEFEF}  &\textbf{~Transfer} & \textbf{Overall}             & \textbf{Vehicle}           & \textbf{Pedestrian}            & \textbf{Cyclist}          \\  \rowcolor[HTML]{EFEFEF} \multirow{-2}{*}{\textbf{3D Object Detector}}
&\textbf{~Paradigm} & \textbf{AP/APH}             &  \textbf{AP/APH}           &  \textbf{AP/APH}            &  \textbf{AP/APH}          \\ \hline\hline
SECOND~\cite{yan2018second}                                    &Scratch &55.08/51.32            & 59.57/59.04	           & 53.00/43.56           & 52.67/51.37     \\
Part-A$^2$-Anchor~\cite{shi2020points}                                    &Scratch &60.39/57.43            & 64.33/63.82           & 54.24/47.11	          & 62.61/61.35
    \\
\hline

PV-RCNN~\cite{shi2020pv}                  &Scratch    & 59.84/56.23   & 64.99/64.38	   & 53.80/45.14	   & 60.72/59.18    \\
GCC-3D (PV-RCNN)~\cite{liang2021exploring}                  &Fine-tuning    & 61.30/58.18$_\text{{(+1.46/+1.95)}}$   & 65.65/65.10   & 55.54/48.02  & 62.72/61.43   \\
Ours (PV-RCNN)                  &Fine-tuning    & {\textbf{62.62/59.28}}$_\textbf{{(+2.78/+3.05)}}$  & 66.04/65.47   & 57.58/49.51   & 64.23/62.86    \\
\hline
CenterPoint~\cite{yin2020center}                  &Scratch    & 63.46/60.95   & 61.81/61.30   & 63.62/57.79& 64.96/63.77    \\
GCC-3D (CenterPoint)~\cite{liang2021exploring}                  &Fine-tuning     & 65.29/62.79$_\text{{(+1.83/+1.84)}}$  & 63.97/63.47   & 64.23/58.47   & 67.68/66.44    \\
Ours (CenterPoint)                 &Fine-tuning     & {\textbf{66.42/63.85}}$_\textbf{{(+2.96/+2.90)}}$   & 64.94/64.42    & 66.13/60.11    & 68.19 	67.01     \\
\hline
CenterPoint-Stage2~\cite{yin2020center}                  &Scratch    & 65.29/62.47  & 64.70/64.11   & 63.26/58.46   & 65.93/64.85   \\
GCC-3D (CenterPoint-Stage2)~\cite{liang2021exploring}                  &Fine-tuning     & 67.29/64.95$_\text{{(+2.00/+2.48)}}$  & 66.45/65.93   & 66.82/61.47   & 68.61/67.46    \\
Ours (CenterPoint-Stage2)                 &Fine-tuning     & \textbf{{68.06/65.69}}$_\textbf{{(+2.77/+3.22)}}$   & 66.98/66.48    & 68.15/62.61   & 69.04/67.97     \\ \hline
\end{tabular}}
\end{table}

\noindent\textbf{Waymo Open Dataset.} Waymo dataset~\cite{sun2020scalability} contains three classes: vehicles, pedestrians, and cyclists. 3D Average Precision (AP) and Average Precision with Heading (APH) are defined as the evaluation metrics for all classes. The AP and APH are based on IoU thresholds of 0.7 for vehicles and 0.5 for pedestrians and cyclists. Two difficulty levels, \ie, LEVEL\_1 and LEVEL\_2 are defined according to the points number in the bounding boxes, where we mainly consider the LEVEL\_2 metric.

 We follow the schedule of OpenPCDet~\cite{openpcdet2020} to fine-tune the detectors on 20\% training samples for 30 epochs.  In particular, we first report the training-from-scratch results of SECOND~\cite{yan2018second}, Part-A$^2$-Anchor~\cite{shi2020points}, CenterPoint~\cite{yin2020center} (VoxelNet version) and PV-RCNN~\cite{shi2020pv}, in terms of GCC3D~\cite{liang2021exploring}. After that, we apply our \textit{ProposalContrast} on two strong baselines, CenterPoint~\cite{yin2020center} and PV-RCNN~\cite{shi2020pv}, to verify our model.
 According to Table~\ref{tb:waymo-D5}, with our self-supervised pre-training, the performances of popular 3D detectors are substantially improved. For PV-RCNN~\cite{shi2020pv}, we improve the model training from scratch by 3.05\% APH, as well as outperform GCC-3D~\cite{liang2021exploring} by 1.10\% APH on average. We further evaluate our pre-trained model on CenterPoint with VoxelNet backbone. The experimental results show that our approach improves 2.9\% and 1.06\% APH, compared with training from scratch and GCC-3D~\cite{liang2021exploring}, respectively. Furthermore, based on the two-stage CenterPoint, our model reaches 65.69\% APH, improving the model trained from scratch by 3.22\%.

We also evaluate our model under a data-efficient 3D object detection setting. To be specific, we split the Waymo training set into two groups, with each group containing 399 scenes ($\sim$80k frames). We first conduct pre-training on one group without using the labels, and then fine-tune the pre-trained model with labels on another group. During fine-tuning, various fractions of training data are uniformly sampled: 1\% (0.8k frames), 10\% (8k frames), 50\% (40k frames) and 100\% (80k frames). Two different backbones, \ie, PointPillars~\cite{lang2019pointpillars} and VoxelNet~\cite{yan2018second} based on CenterPoint, are involved to measure the performance of our pre-trained model. The detection model trained from random initialization is viewed as the baseline. The advantage of our pre-trained model over the baseline is presented in Table~\ref{tb:labeled data}. In essence, our pre-trained model can consistently promote detection performance on both backbones, especially when the labled data is scarce, \ie, improving 8.26\% and 16.95\% APH with 1\% labeled data. Our model also outperforms the baselines under all label settings.


\begin{table}[t]
\caption{{\textbf{Data-efficient 3D object detection on Waymo dataset.} Our \textit{ProposalContrast} consistently improves the performance of modern 3D object detectors, especially when only limited labeled data are available. In each row, we present the results trained from scratch on the top and show the fine-tuning results at the bottom.}}
\label{tb:labeled data}
\centering
\renewcommand\arraystretch{1.00}
\resizebox{0.99\textwidth}{!}{
\begin{tabular}{c||c||c|cccc}
\hline\thickhline
\rowcolor[HTML]{EFEFEF}\textbf{Fine-tuning with}
\cellcolor[HTML]{EFEFEF}                     &&&\multicolumn{4}{c}{\cellcolor[HTML]{EFEFEF}\textbf{3D AP/APH (LEVEL 2)}}  \\
\rowcolor[HTML]{EFEFEF}
\multirow{-2}{*}{\cellcolor[HTML]{EFEFEF}}\textbf{various label ratios} &\multirow{-2}{*}{\textbf{Detector}}
&\multirow{-2}{*}{\textbf{Relative Gain}} & \textbf{Overall}             & \textbf{Vehicle}           &\textbf{ Pedestrian}            & \textbf{Cyclist}         \\ \hline\hline

     &  &   & 23.05/18.08           & 27.15/26.17           & 30.31/18.79          &11.68/9.28       \\
\multirow{2}{*}{1\% ($\sim$ 0.8k frames)}  &\multirow{-2}{*}{PointPillars~\cite{lang2019pointpillars}} &\multirow{-2}{*}{{\textbf{+8.60/+8.26}} } & 31.65/26.34    & 35.88/35.08    & 37.61/25.22    & 21.47/18.73   \\ \cline{2-7}
 &  &   & 20.88/17.83             & 21.95/21.45          & 27.98/20.52         & 12.70/11.53          \\
  & \multirow{-2}{*}{~~VoxelNet~\cite{yan2018second}}&\multirow{-2}{*}{{\textbf{+17.48/+16.95}} } & 38.36/34.78     & 37.60/36.91   & 39.74/31.70   & 37.74/35.73   \\ \hline


   &  &  & 51.75/46.58    & 54.94/54.32   & 54.01/41.53   & 46.31/43.88           \\
\multirow{2}{*}{10\% ($\sim$ 8k frames)}  &\multirow{-2}{*}{PointPillars~\cite{lang2019pointpillars}} &\multirow{-2}{*}{{\textbf{+2.33/+2.85}} } & 54.08/49.43     & 57.54/56.93   & 56.97/45.25   & 47.74/46.1   \\ \cline{2-7}
 &  &   & 54.04/51.24              & 54.37/53.74           & 51.45/45.05          & 56.30/54.93           \\
  & \multirow{-2}{*}{~~VoxelNet~\cite{yan2018second}}&\multirow{-2}{*}{\textbf{{{+4.96/+5.06}}}} &59.00/56.30     & 58.83/58.23   & 57.75/51.75   & 60.42/58.91   \\ \hline


     &  &   & 59.77/55.58            & 61.89/61.32         & 61.89/51.26           & 55.54/54.16          \\
\multirow{2}{*}{50\% ($\sim$ 40k frames)}  &\multirow{-2}{*}{PointPillars~\cite{lang2019pointpillars}} &\multirow{-2}{*}{\textbf{{{+1.06/+1.08}}} } & 60.83/56.66    & 63.01/62.44   & 62.57/52.02   & 56.91/55.53   \\ \cline{2-7}
 &  &   & 63.51/61.05              & 63.18/62.66          & 63.35/57.67           & 63.99/62.82           \\
  & \multirow{-2}{*}{~~VoxelNet\cite{yan2018second}}&\multirow{-2}{*}{\textbf{{{+1.14/+1.05}} }} & 64.65/62.10     & 64.07/63.54   & 64.64/58.71   & 65.24/64.04   \\ \hline

     &  &   & 61.68/57.92           & 63.95/63.39           & 62.91/53.38           & 58.17/57.01           \\
\multirow{2}{*}{100\% ($\sim$ 80k frames)}  &\multirow{-2}{*}{PointPillars~\cite{lang2019pointpillars}} &\multirow{-2}{*}{\textbf{{{+0.54/+0.40}}} } & 62.22/58.32   & 64.05/63.85   & 63.51/53.69   & 58.66/57.41   \\ \cline{2-7}
 &  &   & 64.84/62.29             & 64.38/63.86           & 66.05/60.06          &64.09/62.95          \\
  & \multirow{-2}{*}{~~VoxelNet~\cite{yan2018second}}&\multirow{-2}{*}{\textbf{{{+0.55/+0.61}} }} & 65.39/62.90     & 64.67/64.16   & 66.52/60.65   & 64.97/63.88   \\ \hline

\end{tabular}}
\end{table}

\begin{table}[t]
\caption{{\textbf{3D Object Detection Performance on ONCE validation set.} The improved CenterPoint achieves the best performance among these SOTA detectors.}}
	\label{tab:once}
\centering
\setlength\tabcolsep{8pt}
\renewcommand\arraystretch{1.10}
	\resizebox{0.8\textwidth}{!}
	{%
		\begin{tabular}{c||c| ccc }
			\hline\thickhline
			\rowcolor[HTML]{EFEFEF}&
			&
			\multicolumn{3}{c}{\textbf{Orientation-aware AP}} \\
			\rowcolor[HTML]{EFEFEF}{\multirow{-2}{*}{~~~~~\textbf{Methods}}}&
			\multicolumn{1}{c| }{\multirow{-2}{*}{\textbf{mAP}}}&
			\multicolumn{1}{l}{\textbf{Vehicle}} &
			\multicolumn{1}{l}{\textbf{Pedestrian}} &
			\multicolumn{1}{l}{\textbf{Cyclist}} \\ \hline\hline
			PointPillars~\cite{lang2019pointpillars} & 44.34 & 68.57 &	17.63 &	46.81  \\
			SECOND~\cite{yan2018second} & 51.89 &	71.19 &	26.44 &	58.04 \\
			PV-RCNN~\cite{shi2020pv} & 53.55 &	77.77 &	23.50 &	59.37 \\
			CenterPoint~\cite{shi2020pv} & 60.05 &	66.79 &	49.90 &	63.45 \\
			PointPainting~\cite{vora2020pointpainting} & 57.78 &	66.17 &	44.84 &	62.34 \\
			\hline
			CenterPoint$^{*}$~\cite{yin2020center} & 64.24 &	75.26 &	51.65 &	65.79 \\
			Ours (CenterPoint$^{*}$) & {\textbf{66.24}}$_\textbf{{+2.00}}$ &	\textbf{78.00$_\textbf{{+2.74}}$} &	\textbf{52.56$_\textbf{{+0.91}}$} &	\textbf{68.17$_\textbf{{+2.38}}$} \\  \hline
		\end{tabular}
	}
\end{table}

\noindent\textbf{ONCE Dataset.} ONCE~\cite{mao2021one} is a newly released dataset for 3D object detection in autonomous driving. It involves 5k frames for training or fine-tuning and 3k frames for validation. An orientation-aware AP is introduced to account for objects with opposite orientations. ONCE evaluates 3 categories for 3D object detection: vehicle, pedestrian and cyclist. The official ONCE benchmark provides some results from popular 3D detectors on the validation set. Since our implemented CenterPoint$^{*}$ achieves much better results than the official version, we use it as the baseline. We pre-train the VoxelNet backbone of CenterPoint$^{*}$ on Waymo with our \textit{ProposalContrast} and fine-tune it with the ONCE training set. Then we give a performance comparison with several 3D detectors on the validation set.  As shown in Table~\ref{tab:once}, our model improves the baseline by 2.00\% mAP, achieving better performance on the validation set. This gives another evidence of the transferability of our pre-trained model.

\begin{table}[t]
\caption{{\textbf{Ablation studies} on  different granularities of instance features.} }
\label{tb:pointcontrast}
\centering
\renewcommand\arraystretch{1.10}
\resizebox{0.99\textwidth}{!}{
\centering
\begin{tabular}{c||c|c|ccc}
\hline\thickhline
\rowcolor[HTML]{EFEFEF}
\textbf{Methods}                                                                 & \textbf{Initialization}          &\textbf{Overall (mAP/mAPH) } &\textbf{Vehicle} &\textbf{Pedestrian} &\textbf{Cyclist}\\
\hline\hline
Baseline                    & Random Initialization  &       59.63/56.96  &58.96/58.40 &56.72/50.77 &63.20/61.70\\  \hline\hline
PointContrast~\cite{xie2020pointcontrast}                      & Point-level Pre-train      & 60.32/57.75$_{{(+0.69/+0.79)}}$   &60.47/59.85 &56.67/50.78 &63.82/62.63\\   \cline{1-6}

DepthContrast~\cite{zhang2021self}                      &  Scene-level Pre-train&  60.71/57.96$_{{(+1.08/+1.00)}}$             &60.11/59.62 &58.42/52.27 &63.59/62.00     \\    \hline

\textit{ProposalContrast}                      &  Proposal-level Pre-train&   \textbf{62.75/60.13$_\textbf{{(+3.12/+3.17)}}$}              &61.20/60.64 &60.75/54.83 &66.29/64.93     \\

\hline

\end{tabular}}
\end{table}

\begin{table}[t]
\caption{{\textbf{Ablation studies} on each module of \textit{ProposalContrast}.} }
\label{tb:ablation}
 \renewcommand\arraystretch{1.15}
\resizebox{0.49\textwidth}{!}{
\begin{tabular}{c||c|c|c}
\hline\thickhline
\rowcolor[HTML]{EFEFEF}
\textbf{Modules}                                                                 & \textbf{Aspect}                       & \textbf{Param.}      &\textbf{mAP/mAPH} \\
\hline\hline
Baseline                    & Random Init.  &  - &      59.63/56.96  \\  \hline  
MaxPooling                    & $N$=2048, $r$=1.0  &  - &      62.09/59.32  \\  \hline  
                                                                                             &  \multirow{-0.5}{*}{Proposal}         & 1024    &   62.10/59.47         \\
\multirow{2}{*}{\begin{tabular}[c]{@{}c@{}}Attentive   \\ Proposal    \\ Encoder\ \end{tabular}}                                                                                             &  \multirow{-0.5}{*}{Number ($N$)}        & 2048 &   \textbf{62.75/60.13}                \\
                                                                                             &         & 4096 &   62.54/59.91                  \\ \cline{2-4}

   &    &   0.5   &   62.43/59.81         \\
 &     \multirow{-2}{*}{Spherical Radius}    &   1.0   &  \textbf{62.75/60.13}            \\
 &  \multirow{-2}{*}{ ($r$)  }    &    2.0  &  62.36/59.55         \\   \hline

\end{tabular}}
\quad\hspace{-12pt}
\renewcommand\arraystretch{1.20}
\resizebox{0.49\textwidth}{!}{
\begin{tabular}{c||c|c|c}
\hline\thickhline
\rowcolor[HTML]{EFEFEF}
\textbf{Modules}                                                                 & \textbf{Aspect}                       & \textbf{Param.}      &\textbf{mAP/mAPH}  \\
\hline\hline
Baseline                    & Random Init.  &  - &      59.63/56.96  \\  \hline
IPD task                      & \textit{w/o} ICS  &  -      & 62.17/59.56 \\   \hline

                     &  &    64    & 61.47/58.87 \\
                           \multirow{-0.5}{*}{ICS task}         & \multirow{-2}{*}{$\#$Cluster}  &  128      & 61.77/59.16 \\
                     & \multirow{-2}{*}{(\textit{w/o} IPD) }&  256      & 61.36/58.91 \\ \cline{2-4}
                     & \textit{w/o} SKC &  128      & 57.29/54.52 \\  \hline

IPD + ICS                       &$\#$Cluster&  128   &    \textbf{62.75/60.13}                 \\

   \hline

\end{tabular}}
\end{table}

\subsection{Ablation Study}
\label{sec:ablation}
In this section, we examine our \textit{ProposalContrast} model in depth. We conduct each group experiment by pre-training the VoxelNet backbone on the full Waymo training set in an unsupervised manner, and evaluate the performance by fine-tuning the detector on Waymo 20\% training data. CenterPoint trained from random initialization is viewed as the baseline. 1$\times$ schedule (12 epochs) in~\cite{yin2020center} is used to save computation on the large-scale Waymo.

\noindent\textbf{Comparison to Point-/Scene-level Pre-training.} The main contribution of this work is to propose a proposal-wise pre-training paradigm. To show its merits over previous contrastive learning methods in point cloud, we re-implement the pioneer works like PointContrast~\cite{xie2020pointcontrast}  and DepthContrast~\cite{zhang2021self} based on the VoxelNet backbone of CenterPoint. As seen in Table~\ref{tb:pointcontrast}, \textit{ProposalContrast} achieves much better results when transferring to 3D object detection in large-scale LiDAR point clouds, thanks to the more suitable representation.

\noindent\textbf{Effectiveness of the Proposal Encoding Module.} Next, we ablate the design choices in the proposal encoding module. We first apply a heuristic method (MaxPool) that directly pools the point features inside a proposal and uses liner layers for embedding. As shown in Table~\ref{tb:ablation}, this has obtained improvement over the baseline due to operating on more informative candidate proposals. After being equipped with the attentive proposal encoder, better results are achieved (+0.81 mAPH). This shows the importance of modeling the geometry structures of proposals. After that, we also search the proposal number $N$ and proposal radius $r$. It turns out that $N\!=\!2048$ and $r\!=\!1.0$ m give better results. We infer that more proposals or a larger proposal radius will cause overlaps between neighbor proposals, which may confuse the instance discrimination process.

\noindent\textbf{Effectiveness of the Joint Optimization Module.} We further investigate the effect of the two learning targets in Table~\ref{tb:ablation}. We first consider only the inter-proposal discrimination (IPD) task for self-supervised learning, which improves the baseline by 2.60\% mAPH. Then, we evaluate the pre-training model with only the inter-cluster separation (ICS) task and check the effect of class (cluster) number, \ie, the output dimensions of the predictor. After that, we examine the importance of Sinkhorn-Knopp clustering (SKC). As seen, the performance drops a lot (-2.44 mAPH) without SKC. Finally, the joint learning of the two tasks further improves the performance.


\section{Conclusion}
This paper presented \textit{ProposalContrast}, a proposal-wise pre-training framework for LiDAR-based 3D object detection. Despite the previous works for scene-level or point/voxel-level instance discrimination, we argue that the proposal-level representation is more suitable for 3D object detection in the large-scale LiDAR point cloud, which is not well addressed by previous works. To achieve proposal-wise contrastive learning, we carefully designed a proposal generation module, a region proposal encoding module and a joint optimization module. In particular, the proposal generation module samples dense and diverse spherical proposals with different augmented views. The proposal encoding module abstracts proposal features to model the intrinsic geometry structure of proposals by considering relationships of points inside proposals.
To further build a comprehensive representation, we proposed to jointly optimize an inter-proposal discrimination task and an inter-cluster separation task. Extensive experiments on diverse prevalent 3D object detectors and datasets show the effectiveness of our model. We expect this work will encourage the community to explore the unsupervised pre-training paradigm in driving scenarios.

\noindent\textbf{Acknowledgements.}
This work was partially supported by Zhejiang Lab's International Talent Fund for Young Professionals (ZJ2020GZ023), ARC DECRA DE220101390, FDCT under grant 0015/2019/AKP, and the Start-up Research Grant (SRG) of University of Macau.
\clearpage
%
%
\bibliographystyle{splncs04}
\bibliography{3936.bib}
\end{document}